\documentclass[conference]{IEEEtran}
\usepackage{braket}
\IEEEoverridecommandlockouts

\usepackage{mathpazo}  
\usepackage{inconsolata} 
\usepackage{multirow}
\usepackage{graphicx}
\usepackage{tabularx, colortbl}
\usepackage{array}
\usepackage{amsmath,amssymb,amsfonts}
\usepackage{algorithm}
\usepackage{algpseudocode}
\usepackage{listings}
\usepackage[framemethod=TikZ]{mdframed} 
\usepackage{tcolorbox}
\usepackage{caption}
\usepackage{subcaption}
\usepackage{hyperref}
\usepackage{xcolor}  
\usepackage{cleveref}
\usepackage{lipsum}  

\definecolor{mydarkblue}{rgb}{0.1, 0.1, 0.6}
\hypersetup{
  linkcolor  = mydarkblue,
  citecolor  = orange,
  urlcolor   = cyan,
  colorlinks = true
}

\def\mynameNoSpace{NSP}
\def\myname{NSP }

\usepackage{titlesec}
\titleformat{\section}{\large\bfseries\sffamily}{\thesection}{1em}{}
\titleformat{\subsection}{\bfseries}{\thesubsection}{1em}{}
\titlespacing*{\section}{0pt}{1.5ex plus 1ex minus .2ex}{1ex plus .2ex}

\newtcolorbox{mybox}[1][]{colback=gray!5!white, colframe=gray!75!black, fonttitle=\bfseries, title=#1}

\crefname{section}{Section}{Sections}
\crefname{subsection}{Section}{Sections}

\begin{document}

\title{\huge\textbf{NSP: A Neuro-Symbolic Natural Language Navigational Planner}\\ \vspace{0.4em}}

\author{\IEEEauthorblockN{William English, Dominic Simon, Rickard Ewetz}
\IEEEauthorblockA{\textit{ECE Department}\\ \textit{University of Florida} \\
Gainesville, Florida, 32611, USA \\
\texttt{\{will.english, dominic.simon, rewetz\}@ufl.edu}}

\and
\IEEEauthorblockN{Sumit Kumar Jha}
\IEEEauthorblockA{\textit{Computer Science Department}\\ \textit{Florida International University} \\
Miami, Florida, 33199, USA \\
\texttt{jha@cs.fiu.edu}}
}

\maketitle

\begin{abstract}
\vspace{0.3em}
\begin{tcolorbox}[colback=gray!5!white, colframe=gray!75!black, title=Abstract, boxrule=0.5mm]
Path planners that can interpret free-form natural language instructions hold promise to automate a wide range of robotics applications. These planners simplify user interactions and enable intuitive control over complex semi-autonomous systems. 
While existing symbolic approaches offer guarantees on the correctness and efficiency, they struggle to parse free-form natural language inputs. Conversely, neural approaches based on pre-trained Large Language Models (LLMs) can manage natural language inputs but lack performance guarantees. 
In this paper, we propose a neuro-symbolic framework for path planning from natural language inputs called \mynameNoSpace. The framework leverages the neural reasoning abilities of LLMs to i) craft symbolic representations of the environment and ii) a symbolic path planning algorithm. Next, a solution to the path planning problem is obtained by executing the algorithm on the environment representation. The framework uses a feedback loop from the symbolic execution environment to the neural generation process to self-correct syntax errors and satisfy execution time constraints. 
We evaluate our neuro-symbolic approach using a benchmark suite with 1500 path-planning problems. The experimental evaluation shows that our neuro-symbolic approach produces 90.1\% valid paths that are on average 19-77\% shorter than state-of-the-art neural approaches.


\end{tcolorbox}
\vspace{0.5em}
\end{abstract}


\section{Introduction}
The proliferation of robotics and autonomous systems~\cite{2404.12256} has resulted in the development of numerous complex high-level path planners~\cite{1007.2212, 2207.04429, 2401.03267}. Path planners take an objective and a description of the environment as an input and generate a path planning solution. 
Path planners vary based on the 
 type, format, and completeness expectation of their input and outputs~\cite{2001.02330, 2310.10103, 2205.06301}. It is desirable for planners to be verifiable, or provide some form of correctness guarantee~\cite{1207.2415,2209.09818}. Recent efforts have sought to incorporate Natural Language Processing (NLP) technologies such as Large Language Models (LLMs) in order to create planners that accept natural language as input rather than graphical and spatial input types~\cite{2210.03370}. This development is due to the desire for humans to be able to easily interface with 
 robotic agents using free-form natural language\cite{1701.08756}. Existing solutions to navigational path planning are based on symbolic~\cite{zang-etal-2018-translating, 1810.00663} or neural solution strategies~\cite{1903.01959, 1609.05143}.
 


 Symbolic approaches generate solutions through deterministic rule-based reasoning \cite{2012.13037}, which satisfy correctness guarantees, but are sensitive to non-conformal inputs and are limited by the knowledge embedded in the NL-to-Symbol mapping model~\cite{2402.00854}. Such systems largely rely on specialized resources, such as semantic parsers and lexicons, to interpret free-form natural language instructions, and struggle with the ambiguity of free-form natural language\cite{1701.08756}. They require precise semantic interpretations to map instructions accurately onto a predefined action sequence. For these reasons, symbolic approaches have a limited capacity to work with natural language inputs, and those that do cannot generalize to unseen inputs~\cite{2401.11972}.

Neural approaches produce solutions by mapping inputs directly to actions through learned patterns, which is computationally intensive and typically requires a manually predefined action space~\cite{2012.13037, 2306.06531}. Another line of work in this domain has used LLMs to translate NL instructions into a symbolic expression in a formal system, such as Linear Temporal Logic (LTL) \cite{TUMOVA2016239, 10.1109/IROS51168.2021.9636287}. These approaches, however, require the engineering of a symbolic vocabulary, and other forms of manual labeling. More recently, LMMs have been use to directly solve various forms of reasoning problems~\cite{2305.10037, robotic_neurosymb, text2motion}. This has opened the door to directly solving path planning problems using the innate reasoning abilities of LLMs. The quality of LLM generated solutions can be improved using prompting strategies such as few-shot prompting and chain of thought reasoning~\cite{2305.04091,2201.11903}.


In this paper, we propose a neuro-symbolic framework called \myname for solving path planning problems in free-form natural language. The framework uses a neural LLM to parse the description of the objective and environment. The LLM also ingests the API of a graph library. Next, the LLM produces a symbolic graph representation of the environment and  a path planning algorithm that operates on the graph. The generated code utilizes syntax and API calls from the graph library. Finally, the algorithm is executed on the graph to produce a solution path. A feedback loop is used to allow the LLM to self-correct syntax errors and satisfy execution time constraints. The main contributions of this paper are, as follows:

\begin{itemize}
    \item A neuro-symbolic approach to solving path planning problems in free-form natural language. The approach leverages the advantages of symbolic approaches while circumventing the need for predefined symbolic representations. Both the distance graph and the path planning algorithm are generalized using the LLM. 
    
    
    
    \item The neuro-symbolic feedback loop from the execution environment is able to resolve hallucinations and syntax errors generated by the LLM. The feedback methodology substantially improves the robustness of the natural language to symbolic translation, which is a key challenge for neuro-symbolic approaches. 

    \item The proposed approach is evaluated using a dataset of  1500 natural language path planning scenarios. Compared with state-of-the-art LLM based approaches~\cite{2201.11903}, the \myname framework improves the valid path success rate by up to 76\%. On average, the feedback loop is only required to be executed 1.82 times per input.  

\end{itemize}
\smallskip
The remainder of the paper is organized as follows: related work is reviewed in Section \ref{sec:relatedwork}, a problem formulation is presented in Section \ref{sec:problemformulation}, a description of the \myname{} framework in Section \ref{sec:framework}, a description of our experimental setup  as well as an analysis and discussion of results in Section \ref{sec:evaluation}, and the paper is concluded in Section \ref{sec:conclusion}.

\section{Related Work}
\label{sec:relatedwork}

Large language models (LLMs), such as those from Anthropic and OpenAI, have demonstrated remarkable capabilities in generating human-like text and solving complex tasks, including natural language processing, code generation, and planning. However, these models are known to produce hallucinations, i.e., factually incorrect or contextually inappropriate outputs, which poses a significant risk particularly in safety-critical applications like autonomous systems and planning tasks. Recent work~\cite{milcom2023, icaa2023} has addressed this issue by combining inductive reasoning from LLMs with deductive verification methods such as satisfiability solving. This approach uses formal methods to detect and correct erroneous outputs from LLMs, iteratively guiding them towards producing accurate and verified solutions. By integrating counterexample-guided inductive synthesis, the model is refined through feedback from deductive solvers until a valid solution is obtained. This methodology significantly enhances the trustworthiness and reliability of LLMs in applications where correctness is crucial, such as planning in autonomous systems.

\subsection{Symbolic Methods in Path Planning}
Prior to the broad adaptation of LLMs, NL-based path planners required a method of mapping free-form NL inputs to a symbolic structure. Typically, this mapping is embedded into a model during training and requires formal environmental and behavioral specifications \cite{1810.00663} \cite{wang2019natural}. This specification is necessary both to search a space for optimal solutions and to guarantee the validity of a proposed solution. In these approaches, the NL input of the path planning task is first mapped to a symbolic action space, and the resulting operations are executed deterministically. The scalability of this approach is limited by the up-front cost of engineering the seed lexicons. To adapt to unseen inputs, re-training of the entire model would be necessary, as the planner itself is incapable of reasoning outside the bounds of the action space. Consequently, purely symbolic path planning approaches cannot handle inputs in the form of free form natural language.


\subsection{Large Language Models in Path Planning}
LLMs have been widely adopted as a valuable tool for natural language path planning tasks\cite{2302.05128} \cite{2204.01691}. Their ability to extract dense information from natural language inputs has significantly advanced the state-of-the-art. High-level path planners require some knowledge of their environment to function efficiently, and LLMs have been shown to possess the reasoning capabilities necessary to decode this information from natural language inputs. Furthermore, LLMs themselves produce language-based output which can in turn be used to inform future prompts and contexts, which has been shown to improve reasoning abilities and update the model's knowledge \cite{yamada2024evaluating} \cite{wang2023grammar}
\cite{2212.09561}. 

There are two primary uses of LLMs in path planners. Firstly, they are used as a translator between free-form NL input and an action space. This alleviates the adaptability concerns with regard to symbolic methods, since LLMs are likely to have this knowledge already embedded, although the output of this model is still restricted to the seed lexicon\cite{2306.06531}. Therefore, this approach is only applicable to predefined problem formulations. The second is to perform reasoning in order to create the navigational plan itself. In this case, the model itself provides a path directly. While it is impressive that LLMs can directly solve simple problem instances, the performance rapidly declines when the complexity of the  considered problems are scaled-up. To enable LLMs to solve larger problem instances, the performance can be boosted using prompting strategies. 





\subsection{Approaches to Prompting}
In this section, we discuss how the reasoning capabilities of LLMs can be improved by using prompting methodologies. In particular, we review the concepts of zero-shot prompting, chain-of-thought prompting, and self-consistency. 

\noindent
\textbf{\texttt{Zero\_Shot}}: Zero-shot prompting expects the model to complete the task and return an answer without previous examples or reasoning steps included in the prompt. Zero-shot prompting exemplifies the base abilities of an LLM, relying solely on its embedded knowledge and generalization abilities.
\smallskip

\noindent
\textbf{\texttt{0-CoT}}~\cite{wei2022chain}: Zero-shot chain-of-thought prompting encourages the LLM to articulate intermediate reasoning steps in its response, but does not include reasoning examples. This approach has been shown to improve reasoning capabilities in multiple domains, including finding the shortest path on a graph\cite{2305.10037}. Note that it is difficult to provide few-shot examples when the model must be capable of solving multiple problem types.
\smallskip

\noindent
\textbf{\texttt{SC}~\cite{2203.11171}} is a method of prompting that generates multiple independent reasoning paths and answers. The most common answer is selected as the final answer. This approach attempts to reduce the likelihood of errors by comparing across multiple outputs.

\section{Problem Formulation}
\label{sec:problemformulation}
This paper considers the problem of path planning from natural language. We first define the natural language path planning problem. Next, we provide a small example of a problem instance and the expected output. 

\noindent
\textbf{Problem Definition:} The input is described in natural language and consists of i) a description of the environment, ii) the objective of the path planning problem, and iii) any constraints in the environment. A path planner is expected to take the input specification and generate a solution path through the environment.  

Formally, we define the path planning problem using a string triplet $\mathcal{D} = \mathcal{(E,S,C)}$, where $\mathcal{E}$ is a string in NL that contains a description of all locations in an environment, as well as the spatial relationships between these locations. The description may include specific distances between pairs of locations or default to unit distances. The description is constructed based on an underlying ground truth graph $G_T=(V_T,E_T)$, where the vertices and the weighted edges capture the locations and the distances between connected spatial locations, respectively. 
$\mathcal{S}$ is an NL string containing the objective of the  path planning problem. The objective may include both shortest path problems and traveling salesman problems. $\mathcal{C}$ is an NL string that contains all constraints on $\mathcal{S}$. The constraints include locations and spatial connections that may not be utilized in the solution path. Most importantly, the options and syntax of each of the individual strings and their concatenations should be treated as unknown to the path planner. The path planner is expected to generate a solution path $\mathcal{P}$, consisting of a set of vertices $\{v_1, \dots, v_n\}$. For the path to be valid, all vertices are required to belong to $V_T$ and all pairs of adjacent vertices $(v_i, v_{i+i}) \in \mathcal{P}$ are required to exist in $E_T$. If the solution path is valid, the path length is compared with the known optimal lower bound from the ground truth graph $G_T$.








\begin{figure*}[t]  
\centering
\begin{minipage}[b]{\textwidth}
    \centering
    \includegraphics[width=15cm]{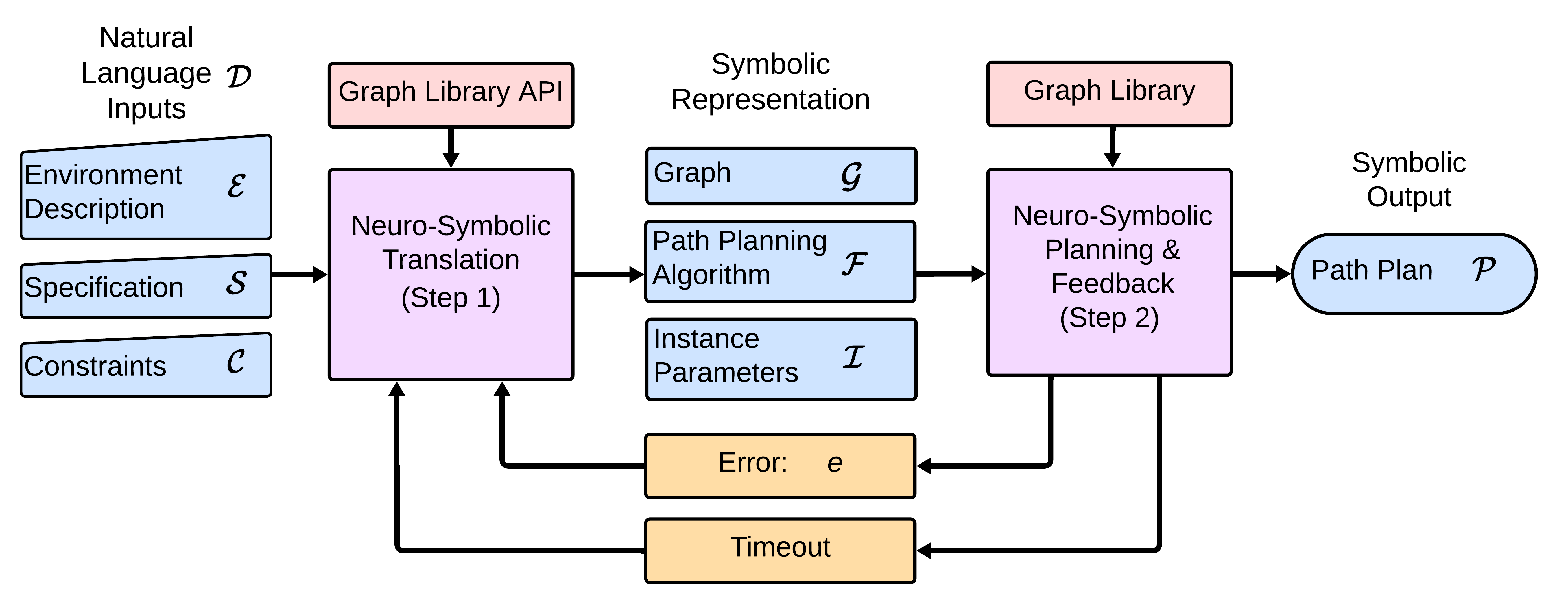}
    \caption{Overview of the \myname{} framework. In Step 1, an arbitrary free-form path planning problem $\mathcal{D} = \braket{\mathcal{E+S+C}}$ is mapped to a symbolic representation consisting of graph, a path planning algorithm, and the parameters accepted by the algorithm. These objects are passed to Step 2 to be interpreted as $\mathcal{F(G, I)}$. In the event that this triggers a timeout or interpreter error, the feedback loop is triggered to improve the plan (repeat Step 1). If the feedback loop is not triggered, the symbolic path plan is returned.}
    \label{fig:1}
\end{minipage}
\end{figure*}

\noindent
\textbf{Sample Problems:} We provide two small sample problems considered by our problem formulation below.

``\textit{There is a house with Room1, Room2, Room3, and Room4. Room1 is connected to Room2 and Room3, Room2 is connected to Room3 and Room4, and Room3 is connected to Room4. Start in Room1 and move to Room2. Do not pass through Room4.}'' The expected output from this problem instance would be ``\textit{[Room1, Room3, Room2]}".

``\textit{There is a house with Room1, Room2, Room3, and Room4. Room1 is connected to  Room2 with a weight of $3$ and Room4 with a weight of $5$, Room2 is connected to Room4 with a weight of $3$ and Room3 with a weight of $2$, and Room3 is connected to Room4 with a weight of $1$. Find a path that begins and ends in Room1 that passes through each room at least once.}'' The expected output from this problem instance would be ``\textit{[Room1, Room4, Room3, Room2, Room1]}". 


\section{The \myname{} Framework}
\label{sec:framework}


In this section, we describe the proposed \myname framework, which is a neuro-symbolic approach to solving path planning problems in free-form natural language. The input to the \myname{} framework is the free-form natural language description of the path planning problem, $\mathcal{D} = \braket{\mathcal{E+S+C}}$. The output is a solution to the problem, the path $\mathcal{P}$. The neuro-symbolic translation involves using an LLM to translate the natural language input into symbolic representations, including a graph and parameters $\braket{\mathcal{G, I}}$, an algorithm $\mathcal{F}$ to solve the path planning problem, and a function call  $\mathcal{F(G, I)}$. The LLM ingests the API of a graph library to provide tools for the LLM to utilize in its code generation. 
An overview of the framework is provided in Figure~\ref{fig:1}. The details of the neuro-symbolic translation are provided in Section~\ref{subsec:4a}. The neuro-symbolic planning and feedback step involves executing the function call generated in the previous step. If the symbolic path planning algorithm can be executed as expected, this step directly produces a solution $ \mathcal{P}$ to the path planning problem. If a compilation error is encountered during the execution, or if the execution time exceeds a limit, the (neural) feedback is provided back to the neuro-symbolic translation in the form of natural language. The details of the neuro-symbolic planning and feedback are given in Section~\ref{subsec:4b}. The flow is illustraited with a case study in Section~\ref{subsec:4c}.




\begin{figure}
\noindent\fbox{%
    \parbox{\linewidth}{%
    prompt = ``The path planning problem is:
    \{\textbf{problem description $\mathcal{D}$}\}
    \smallskip
    
    \smallskip
    Write a Python function '\textbf{create\_graph()}' that generates a distance graph using the NetworkX library based on the path planning problem. The function should return a NetworkX weighted graph object.
    \smallskip
    
    \smallskip
    Additionally, write another function '\textbf{solve\_problem(graph, args)}' that solves the path planning problem in the form of a node traversal order list.
    Your response should include the complete function code and a definition of args, which is an array containing any parameters you may need for the solution function, in your response. Do not return any incomplete functions.
    \smallskip
    
    \smallskip
    The available libraries are networkx and itertools.
    If this problem is similar to another problem with a known efficient solution, use it in your implementation. 
    \smallskip
    
    \smallskip
    {\{\textbf{graph library API}\}}
    }%
}
    \caption{The prompt template used within the neuro-symbolic translation is shown above. The template captures the problem description, the expect outputs, and the API of  graph library.}
    \label{fig:promptstruct}
    \vspace{-12pt}
\end{figure}

\subsection{Neuro-Symbolic Translation}
\label{subsec:4a}
The first step in the \myname framework is the neuro-symbolic translation. The goal of this step is to translate the natural language inputs into an algorithm $\mathcal{F}$ that takes a graph $\mathcal{G}$ and  parameters $\mathcal{I}$ as the input. By executing $\mathcal{F(G,I)}$, the algorithm should generate a solution path $\mathcal{P}$.

The translation of the natural language inputs to the symbolic intermediate representations is performed using an LLM. While LLMs are capable of generating high quality code from scratch, the desired task is challenging to be performed zero-shot. To improve the quality of the code generation, we leverage a combination of i) access to graph libraries and ii) clever prompting strategies. While the path planner does not know the exact problem that it will be requested to solve, most planning problem can be modeled and solved using distance graphs. Therefore, we provide the API of a graph library in the prompt to the LLM. Our prompting strategy takes inspiration from prior work that has established prompt patterns that are most successful in eliciting correct code generation\cite{2302.11382}. Specifically, we provide the names and parameters of the functions, a description of their purpose, what libraries are available, and the names of relevant variables that need to be defined. 

Our developed prompt template for solving the path planning problems is shown in Figure~\ref{fig:promptstruct}. The  template delineates requirements for the type of code that will be output by the model. The features of the input that are not predefined - that is, the problem description $\braket{\mathcal{D}=(E, S, C)}$ - contain all the information required by the model to successfully map our inputs to a partial solution $\braket{\mathcal{F(G, I)}}$. The model is asked to write two functions: a function that returns a graph representation of the environment $\mathcal{G}$, and a function $\mathcal{F}$ that accepts $\braket{\mathcal{G, I}}$ as input and returns $\mathcal{P}$. In addition to these functions, this step also stores $\braket{\mathcal{I}}$ in local variables that are necessary to run the path-finding algorithm. Information about the Graph API is provided to the model in the prompt. This information consists of functions headers for API calls relevant to path-finding problems on graphs.

\begin{figure*}[ht!]
    \centering
    \includegraphics[width=\textwidth]{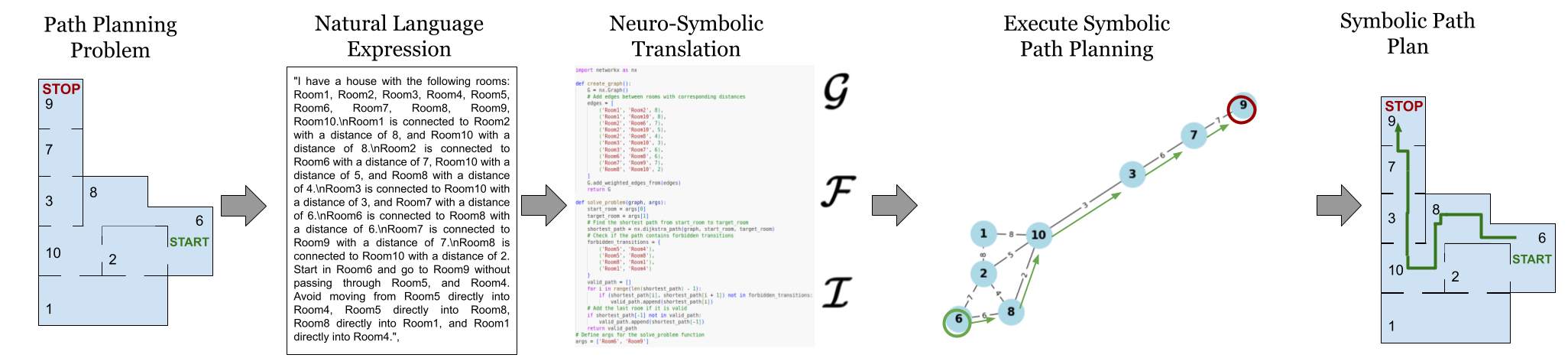}
    \caption{The flow of information in the \myname{} framework. The Neuro-Symbolic Translation can be found in Section \ref{subsec:4c}.}
    \label{fig:representations}
    \vspace{-12pt}
\end{figure*}

\subsection{Neuro-Symbolic Planning \& Feedback}
\label{subsec:4b}

The second step of the \myname{} framework is the neuro-symbolic feedback \& planning step. The inputs are the intermediate components $\braket{\mathcal{G, I, F}}$ generated in the previous step. In this step, we execute the algorithm $\mathcal{F}$ on the arguments $\braket{\mathcal{G, I}}$. The result is one of three types of feedback: an interpreter error, a timeout error, or a solution path $\mathcal{P}$. If a solution path is generated, the path $\mathcal{P}$ is returned as the solution to the considered problem description $\mathcal{D}$. Otherwise, we trigger a feedback loop to handle either the interpretation error, or the timeout error. If the feedback loop is invoked more than $m=5$ times, we report that \myname is unable to generate a valid solution path.  


The use of a feedback loop is motivated by the fact that an LLM by itself can not reliably write flawless code in one attempt. 
Syntactical or logical errors may be present, and so before we finalize our path plan, we run the algorithm with the compiler or interpreter of the programming language. In our implementation, we leverage the Python programming language and the Python interpreter. In the case that the interpreter detects an error in the code or the code does not execute within the specified time, we append the respective errors to the input and repeat step 1, i.e., the neuro-symbolic translation in Section~\ref{subsec:4a}. 
Specifically, we append the error template shown in Figure~\ref{fig:error}. If the interpreter did not produce an error, the path is considered verified and correct.

\begin{figure}[H]
\noindent\fbox{%
    \parbox{\linewidth}{%
        if error\_message:
        \smallskip
    
        \smallskip
        \{\textbf{prompt}\} += ``An error occurred with the previous response: 
                  \textbf{\{function\_code\}}
                  The error message was: \textbf{\{error\_message\}} Please correct the response."
    \smallskip
    
    *if code timed out, error\_message = "The code took too long to execute. Increase the time efficiency or approximate the solution."
        }%
}
    \caption{An error message detected by the interpreter is appended to the prompt prior to repeating neuro-symbolic translation (step 1).}
    \label{fig:error}
\end{figure}

The error template is designed to handle two types of errors, both syntax errors and timeout errors. The first part of the template is geared to capture syntax errors. We observe that the generated code may contain errors such as \texttt{KeyError}, \texttt{NameError}, \texttt{ValueError}, \texttt{AttributeError}, etc. A detailed analysis of these errors and the corrections made after the feedback loop is provided in Section~\ref{sec:evaluation}. Most of them step from querying undefined objects or non-existent properties. When the LLM is provided this information process as feedback, it is capable of correcting itself in most cases and generating functional code in the next iteration. This feedback loop is one of the key features that allow the \myname framework to successfully generate a symbolic representation from the natural inputs, which is a key challenge to neuro-symbolic approaches for path planning problems.
The second type of feedback message that can occur is a timeout error. We have specified a time limit $t$ for the maximum execution time of the function call $\mathcal{F=(G,I)}$. We set the maximum time to one minute in our implementation. If the code has not converged to a solution path within the time limit, we ask the LLM to generate more efficient code, or solve the problem approximately. This type of feedback loop avoids infinite loops to break the framework, or exact solutions being employed to solve large scale problems that are NP hard, i.e., for example the TSP problem.

\subsection{Case Study on a Small Example}
\label{subsec:4c}
\begin{figure}
    \includegraphics[width=0.5\textwidth]{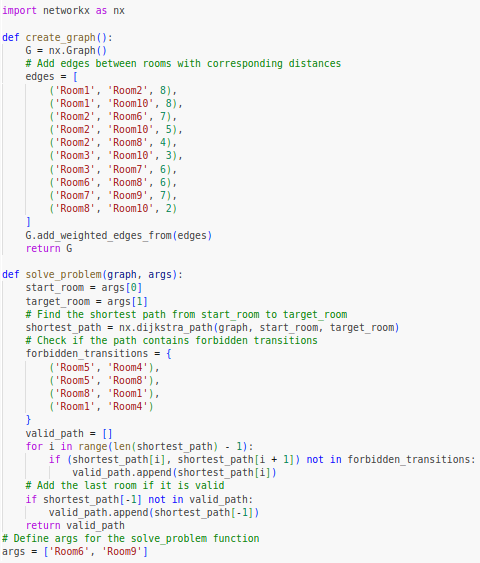}    
\caption{Python code resulting from example in Section \ref{subsec:4c}. Segment 1: Returns $\mathcal{G}$, a graph that is perfectly consistent with the natural language description of the house contained in the input. Segment 2: Takes $\braket{\mathcal{G, A}}$ as input and returns the symbolic path plan $\mathcal{P}$ as output. Segment 3:  $\braket{\mathcal{A}}$, the parameters defined by the model, which are passed to the function defined in segment 2. Segment 4: The code contained in segments 1 through 3 is executed, and its output is stored in the variable \textit{generated\_path} = $\mathcal{P}$ .}
\label{fig:execution}
\end{figure}

In this section, we provide an example of the \myname framework solving a simple path planning problem. An overview of the  solution process is illustrated in Figure~\ref{fig:1}. Some detailed generated code is shown in Figure~\ref{fig:execution}. The case study is performed using the experimental setup described in Section~\ref{fig:execution}. 

The case study considers the following problem description $\mathcal{D}$ ``I have a house with the following rooms: Room1, Room2, Room3, Room4, Room5, Room6, Room7, Room8, Room9, and Room10. Room1 is connected to Room2 with a distance of 8, and Room10 with a distance of 8. Room2 is connected to Room6 with a distance of 7, Room10 with a distance of 5, and Room8 with a distance of 4. Room3 is connected to Room10 with a distance of 3, and Room7 with a distance of 6. Room6 is connected to Room8 with a distance of 6. Room7 is connected to Room9 with a distance of 7. Room8 is connected to Room10 with a distance of 2. Start in Room6 and go to Room9 without passing through Room5, and Room4. Avoid moving from Room5 directly into Room4, Room5 directly into Room8, Room8 directly into Room1, and Room1 directly into Room4."

Below are three Python code segments produced in response to this input.
The first code segment is the graph object that is created using the NetworkX graph package. The second code snippet is the function call that is used to solve the path planning problem. The algorithm also contains a list of edges that are not allowed to be traversed. However, it can be observed that the algorithm performs this as a post-check instead of removing the edges from the graph ahead of time. The last code snippet are the arguments of to the function call, i.e., the start and end room. 

\section{Experimental Evaluation}
\label{sec:evaluation}
The experimental evaluation is broken into the following sections: construction of the evaluation dataset is explained in Section~\ref{sec:dataset_construction}, experimental setup is given in Section~\ref{sec:setup}, results on the Shortest Path and Traveling Salesman problems are discussed in Section~\ref{subsec:shortestpathtask} and Section~\ref{subsec:tsptask}, respectively, error analysis in Section~\ref{subsec:erroranalysis}, and a general discussion in Section~\ref{subsec:evalsummary}.


\subsection{Dataset Construction}
\label{sec:dataset_construction}
The evaluation dataset is composed of 1500 randomly generated navigational path planning scenarios. Each scenario is classified by the parameters $num\_rooms \in [5, 10, 15, 20, 25]$ and $graph\_type \in [weighted, unweighted]$. These scenarios are further divided into Shortest Path problems and Traveling Salesman problems. Shortest Path problems are divided into $constrained \in [true, false]$ denoting whether the path planning problem $\mathcal{D}$ includes any constraints $\mathcal{C}$. There are $30$ unique combinations of the scenario parameters and $50$ scenarios for each combination of scenario parameters, resulting in a total of 1500 scenarios. In addition to the parameters of each scenario, each entry includes a string description of the environment $S$, a string description of the path planning problem $P$, a graph isomorphic with the house in the description $\mathcal{G} = (V, E, w)$, where $V = \{v_1, v_2, ..., v_n\}$, $E = \{(v_i, v_j) | v_i, v_j \in V, i \neq j\}$, and $w \in \mathbb{R}$, and the optimal solution path $\mathcal{P} = (v_1, v_2, ... v_k)$ for this scenario.


The Shortest Path graphs are constructed by initially creating a complete graph. The number of nodes in the given graph is determined by the $num\_rooms$ parameter. Then, edges between the nodes are randomly removed until $40\%$ of the edges remain. Finally, two nodes are randomly chosen as the start and end nodes. The shortest path between these nodes becomes the optimal path $\mathcal{P}$. The graph is algorithmically traversed to build the natural language descriptions of the pairs $(\mathcal{G}, \mathcal{P})$. The Traveling Salesman graphs are built in an inverse manner. The nodes are initially fully unconnected, containing no edges. Two nodes are chosen as the start and stop nodes, and an optimal path $\mathcal{P}$ is generated. Edges are then randomly added to the graph until $40\%$ of the possible edges exist.


\subsection{Evaluation Setup}
\label{sec:setup}
We define a set of metrics to evaluate the navigation performance of the proposed method for each scenario type. These metrics include Success Rate, Optimal Path Rate, Path Efficiency Rate, and Inference Time. \textbf{Success Rate} is defined as $\mathcal{P}_C \div \mathcal{P}_T$, the number of path solutions that successfully reach the end node $\mathcal{P}_C$ divided by the total number of path solutions generated $\mathcal{P}_T$. \textbf{Optimal Path Rate} is defined as $\mathcal{P}_O \div \mathcal{P}_C$, where $\mathcal{P}_O$ is the number of solution paths that reach end node using the shortest path. \textbf{Path Efficiency Rate} is defined as $\sum_{}^{}len(\mathcal{P}_O) \div \sum_{}^{}len(\mathcal{P}_C)$, the sum of the edge lengths $len()$ of optimal paths $\mathcal{P}_O$ divided by the sum of the edge lengths of the generated paths $\mathcal{P}_C$. A perfect performance is indicated by a score of 100\% in each of these metrics. \textbf{Inference Time} is the time it takes for the model API call to return the output. \textbf{Attempts} is also included specifically for \myname{} to show the average number of feedback loop iterations per trial.

We evaluate our approach against four baselines. These include zero-shot prompting (\texttt{Zero\_Shot}), zero-shot chain-of-thought prompting (\texttt{0-CoT}), zero-shot + self-consistency prompting (\texttt{Zero\_Shot+SC}), and zero-shot chain-of-thought + self-consistency prompting (\texttt{0-CoT+SC}). Each of these approaches are described in Section \ref{sec:relatedwork}. We use these prompting approaches to demonstrate the performance gains achieved by \myname{} over other techniques. 

\texttt{GPT-4o-mini} is used as the main LLM during evaluation. At the time of writing, \texttt{GPT-4o-mini} is the newest LLM from OpenAI, providing the best price-per-token to processing power tradeoff. 


Experimental evaluations were performed on a desktop with an Intel Core i7-137000 (16 cores, 5.2 GHz) CPU, an NVIDIA RTX A4000 (16 GB) GPU, and 32 GB of RAM.

\subsection{Shortest Path Problem}
\label{subsec:shortestpathtask}

\begin{table*}[t!]
\vspace{-6pt}
\caption{Experimental results for the Shortest Path Problem using houses with various number of rooms.}
\label{tab:sp}
\centering
\begin{tabular}{|l|c|r|r|r|r|r|}
\hline
\multicolumn{1}{|l|}{Method} & \multicolumn{1}{c|}{Num} & \multicolumn{1}{c|}{Success} & \multicolumn{1}{c|}{Optimal Path} & \multicolumn{1}{c|}{Path Efficiency} & \multicolumn{1}{c|}{Feedback} & \multicolumn{1}{c|}{Inference}\\

\multicolumn{1}{|c|}{} & \multicolumn{1}{c|}{Rooms} & \multicolumn{1}{c|}{Rate} & \multicolumn{1}{c|}{Rate} & \multicolumn{1}{c|}{Rate} & \multicolumn{1}{c|}{Calls} & \multicolumn{1}{c|}{Times} \\
\multicolumn{1}{|c|}{} & \multicolumn{1}{c|}{(num)} & \multicolumn{1}{c|}{(\%)} & \multicolumn{1}{c|}{(\%)} & \multicolumn{1}{c|}{(\%)} & \multicolumn{1}{c|}{(Avg)} & \multicolumn{1}{c|}{(s)} \\
\hline
\texttt{0-CoT} & 5 & 93 & 81 & 93 & - & 5.49\\
\texttt{0-CoT+SC}& 5 & 99 & 91 & 97 & - & 27.59\\
\texttt{Zero\_Shot} & 5 & 98 & 80 & 90 & - & 2.67\\
\texttt{Zero\_Shot+SC} & 5 & 97 & 88 & 97 & - & 16.60\\
\texttt{NSP} (Ours)  & 5 & 98 & 98 & 100 & 1.0 & 3.57\\
\hline
\texttt{0-CoT}& 10 & 79 & 44 & 84 & - & 8.30\\
\texttt{0-CoT+SC}& 10 & 84 & 45 & 89 & - & 39.21\\
\texttt{Zero\_Shot} & 10 & 81 & 54 & 83 & - & 5.14\\
\texttt{Zero\_Shot+SC} & 10 & 83 & 51 & 89 & - & 30.49\\
\texttt{NSP} (Ours) & 10 & 98 & 97 & 100 & 1.0 & 5.00\\
\hline
\texttt{0-CoT}& 15 & 73 & 14 & 85 & - & 10.11\\
\texttt{0-CoT+SC}& 15 & 73 & 19 & 87 & - & 46.00\\
\texttt{Zero\_Shot} & 15 & 48 & 19 & 78 & - & 6.84\\
\texttt{Zero\_Shot+SC} & 15 & 53 & 23 & 84 & - & 36.70\\
\texttt{NSP} (Ours) & 15 & 99 & 97 & 100 & 1.1 & 6.72\\
\hline
\texttt{0-CoT}& 20 & 70 & 6 & 82 & - & 11.14\\
\texttt{0-CoT+SC}& 20 & 78 & 2 & 86 & - & 50.41\\
\texttt{Zero\_Shot} & 20 & 30 & 7 & 76 & - & 7.50\\
\texttt{Zero\_Shot+SC} & 20 & 21 & 4 & 84 & - & 34.88\\
\texttt{NSP}(Ours) & 20 & 98 & 96 & 99 & 1.0 & 9.16\\
\hline
\texttt{0-CoT}& 25 & 74 & 0 & 83 & - & 11.82\\
\texttt{0-CoT+SC}& 25 & 77 & 1 & 77 & - & 51.05\\
\texttt{Zero\_Shot} & 25 & 17 & 1 & 76 & - & 9.40\\
\texttt{Zero\_Shot+SC} & 25 & 14 & 2 & 80 & - & 35.78\\
\texttt{NSP} (Ours) & 25 & 100 & 97 & 99 & 1.0 & 10.85\\
\hline
\end{tabular}
\end{table*}

In this section, we discuss the results of evaluations on each approach for the Shortest Path problem. The Shortest Path problem consists of finding the shortest path between two locations and the constraints may include forbidden rooms and/or forbidden traversals between rooms.

Table~\ref{tab:sp} shows the performance of the evaluated approaches across path planning scenarios containing $5$, $10$, $15$, $20$, and $25$ rooms, with each room set containing $200$ scenarios. For the 5-room path planning scenarios, all approaches achieve high success rates, with \texttt{0-CoT} at $93\%$, \texttt{0-CoT+SC} at $99\%$, \texttt{Zero\_Shot} at $98\%$, \texttt{Zero\_Shot+SC} at $97\%$, and \texttt{NSP} at $98\%$. As the number of rooms in the path planning scenario increases to $15$ rooms, the success rates for baseline all approaches decreases by an average of $35\%$ while \texttt{NSP} continues to achieve a high success rate of $99\%$. At the 25-room scenarios, the \texttt{0-CoT} baselines maintain a $76\%$ average accuracy while the \texttt{Zero\_Shot} baselines fail almost completely, achieving only a $16\%$ average accuracy. In contrast, \texttt{NSP} contains to maintain a high success rate of $100\%$ on the 25-room scenarios, showing that \texttt{NSP} is able to reliably solve shortest path planning problems of varying sizes significantly more reliably than existing baselines. 

\texttt{NSP} drastically outperforms the baseline approaches on the Optimal Path Rate metric. For 5-room scenarios, \texttt{0-CoT} achieves an $81\%$ optimal path rate, \texttt{0-CoT+SC} achieves a $91\%$ optimal path rate, \texttt{Zero\_Shot} achieves an $80\%$ optimal path rate, \texttt{Zero\_Shot+SC} achieves an $88\%$ optimal path rate, and \texttt{NSP} achieves an $98\%$ optimal path rate. This means that \texttt{NSP} is generating an optimal path for $98\%$ of the solutions that are valid paths. The optimal path rate for the baselines continues to decrease up to the 25-room scenario. \texttt{NSP} still achieves a high optimal path rate of $97\%$ for the 25-room scenario.\texttt{NSP} not only reliably solves path planning problems, it solves those problems optimally nearly every time.

A similar trend can be seen with the Path Efficiency Rate. From the 5-room scenario to the 25-rooms scenario, \texttt{0-CoT} efficiency rate drops by 10\%, \texttt{0-CoT+SC} efficiency rate drops by 20\%, \texttt{Zero\_Shot} efficiency rate drops by 14\%, and \texttt{Zero\_Shot+SC} efficiency rate drops by 17\%. Meanwhile, the efficiency rate of \texttt{NSP} only drops from $100\%$ at 5-rooms to $99\%$ at 25-rooms. The discrepancy in achieving a $100\%$ path efficiency rate while achieving below a $100\%$ optimal path rate can be explained by a very low number of sub-optimal edge traversals coupled with how we have rounded the values in the table. \texttt{NSP} is only traversing a few sub-optimal edges, so the percentage lost on the path efficiency rate is low enough that the percentage gets rounded up. These results further reinforce the idea that \texttt{NSP} is adept at finding optimal paths and also shows that even when \texttt{NSP} generates a sub-optimal path, the number of sub-optimal edges traversed is not high. 

The Feedback Calls column measures the average number of times \texttt{NSP} triggers its feedback loop. The feedback loop is triggered an average of $1.065$ times across all five different numbers of room scenarios. The average number of feedback calls is $1.07$ for the 25-room scenarios, which \texttt{NSP} achieves a $100\%$ success rate on. This supports the idea that the feedback loop is actively beneficial to \texttt{NSP}, allowing it more chances retry scenarios it has failed that it might usually correctly solve. This is the explanation for why our neuro-symbolic approach can provide good results, despite the challenging free-form natural language inputs. 

The final column in Table~\ref{tab:sp} is the average Inference Time in seconds for each approach. \texttt{NSP} is the fastest approach up to the 15-room scenarios, but is becomes slower than \texttt{Zero\_Shot} for the 20-room and 25-room scenarios. A factor in the increased inference time is the existence of the feedback loop in \texttt{NSP}, which causes the method to rerun possibly multiple times. This is not so much of an issue at lower room counts, but at the higher room counts, the increase in run time becomes more visible. This is the main trade-off of \texttt{NSP}, as increasing or decreasing the number of allowed feedback loops will conversely increase or decrease the total inference time. 



\begin{table*}
\vspace{-12pt}
\centering
\caption{Experimental results for the Traveling Sales Man Problem using houses with various number of rooms.}
\begin{tabular}{|l|c|r|r|r|r|r|}
\hline
\multicolumn{1}{|l|}{Method} & \multicolumn{1}{c|}{Num} & \multicolumn{1}{c|}{Success} & \multicolumn{1}{c|}{Optimal Path} & \multicolumn{1}{c|}{Path Efficiency} & \multicolumn{1}{c|}{Feedback} & \multicolumn{1}{c|}{Inference}\\
\multicolumn{1}{|c|}{} & \multicolumn{1}{c|}{Rooms} & \multicolumn{1}{c|}{Rate} & \multicolumn{1}{c|}{Rate} & \multicolumn{1}{c|}{Rate} & \multicolumn{1}{c|}{Calls} & \multicolumn{1}{c|}{Times} \\
\multicolumn{1}{|c|}{} & \multicolumn{1}{c|}{(num)} & \multicolumn{1}{c|}{(\%)} & \multicolumn{1}{c|}{(\%)} & \multicolumn{1}{c|}{(\%)} & \multicolumn{1}{c|}{(Avg)} & \multicolumn{1}{c|}{(s)} \\
\hline
\texttt{0-CoT}  & 5 & 53 & 53 & 67 & - & 11.98 \\
\texttt{0-CoT+SC} & 5 & 72 & 42 & 81 & - & 50.53 \\
\texttt{Zero\_Shot} & 5 & 62 & 62 & 63 & - & 6.32 \\
\texttt{Zero\_Shot+SC} & 5 & 73 & 73 & 73 & - & 28.72 \\
\texttt{NSP} (Ours) & 5 & 76 & 53 & 87 & 2.9 & 4.94 \\
\hline
\texttt{0-CoT}  & 10 & 21 & 21 & 24 & - & 12.44 \\
\texttt{0-CoT+SC} & 10 & 28 & 15 & 31 & - & 63.19 \\
\texttt{Zero\_Shot} & 10 & 28 & 28 & 30 & - & 6.91 \\
\texttt{Zero\_Shot+SC} & 10 & 33 & 33 & 33 & - & 27.90 \\
\texttt{NSP} (Ours) & 10 & 79 & 19 & 75 & 3.0 & 6.13 \\
\hline
\texttt{0-CoT}  & 15 & 11 & 11 & 12 & - & 12.54 \\
\texttt{0-CoT+SC} & 15 & 9 & 6 & 10 & - & 63.39 \\
\texttt{Zero\_Shot} & 15 & 7 & 7 & 9 & - & 6.26 \\
\texttt{Zero\_Shot+SC} & 15 & 8 & 6 & 8 & - & 26.44 \\
\texttt{NSP} (Ours) & 15 & 83 & 1 & 75 & 3.0 & 8.93 \\
\hline
\texttt{0-CoT}  & 20 & 2 & 2 & 2 & - & 12.58 \\
\texttt{0-CoT+SC} & 20 & 2 & 2 & 2 & - & 66.95 \\
\texttt{Zero\_Shot} & 20 & 1 & 1 & 1 & - & 5.42 \\
\texttt{Zero\_Shot+SC} & 20 & 2 & 1 & 2 & - & 27.02 \\
\texttt{NSP} (Ours) & 20 & 93 & 1 & 76 & 2.9 & 15.88 \\
\hline
\texttt{0-CoT}  & 25 & 0 & 0 & 0 & - & 13.15 \\
\texttt{0-CoT+SC} & 25 & 1 & 1 & 1 & - & 62.20 \\
\texttt{Zero\_Shot} & 25 & 0 & 0 & 0 & - & 5.93 \\
\texttt{Zero\_Shot+SC} & 25 & 0 & 0 & 0 & - & 28.24 \\
\texttt{NSP} (Ours) & 25 & 77 & 0 & 75 & 3.2 & 17.78 \\
\hline
\end{tabular}
\label{tab:tsp}
\vspace{-6pt}
\end{table*}


\subsection{Traveling Salesman Problem (TSP)}
\label{subsec:tsptask}
In this section we discuss results of our evaluation of each approach on the Traveling Salesman problem. The Traveling Salesman problem consists of finding a path that visits each location in an environment exactly once, beginning and ending in a specified location.
A successful TSP path plan must at least visit every location in the input. We have verified that every TSP trial in our evaluation dataset contains a cycle that visits each node. 

Table~\ref{tab:tsp} shows the performance of the evaluated approaches across path planning problem scenarios containing 5, 10, 15, 20, and 25 rooms, where each different room set contains $100$ scenarios. For 5-room scenarios, the different methods achieve a success rate in the 50\% to 76\% range, with our proposed \texttt{NSP} method achieving the highest success rate of 76\%. As the number of rooms increases, the success rate of the baselines approaches rapidly decreases, reaching a $0\%$ or nearly $0\%$ success rate at the 25-room scenarios. \texttt{NSP} outperforms the baselines, keeping a success rate of over $75\%$ across all of the different room counts. \texttt{NSP} achieves an optimal path rate of $53\%$ on the 5-room scenarios, but its optimal path rate drops to $0\%$ with the rest of the baselines by the 25-room problem. However, the low optimal path rate of the approaches is not surprising as the traveling salesman problem is an NP-complete problem that is not easy to solve using intuition except for small problem instances. 

We also observe that the \myname{} framework utilizes $2.574$ feedback calls per problem instance to generate a valid solution. This indicates that it is significantly harder for the LLM to generate the code for the TSP problem than the shortest path problem, which only used an average of $1.065$ feedback loops. This may stem from the fact that the LLM has seen more examples of shortest path problems than TSP problems in the training data as shortest path problems are easier to solve and therefore more likely to appear in training data. This increase in feedback calls is also a factor in the increased inference times. Still, \texttt{NSP} is able to consistently perform the second best compared to the baselines, with \texttt{Zero\_Shot} being the only baselines to run faster than \texttt{NSP}. To echo the discussion around inference time in Section~\ref{subsec:shortestpathtask}, we believe that if the number of allowed feedback loops were decreased, \texttt{NSP} would have faster inference times at the cost of lower success rates. 

\subsection{Error Analysis}
\label{subsec:erroranalysis}
In this section we review the most common errors caught by the \myname{} feedback loop. At the conclusion of the evaluation, there were a total of 57 unaccounted errors remaining, i.e, 57/1500 trials were failed because errors were not resolved after the maximum number of feedback iterations was reached. Among these 57 occurences are 17 \texttt{KeyErrors}, 13 \texttt{TypeErrors}, 10 \texttt{NameErrors}, 6 \texttt{NetworkXErrors}, 4 \texttt{OtherErrors}, 3 \texttt{ValueErrors}, 2 \texttt{IndexErrors}, 1 \texttt{UnboundLocalError}, and 1 \texttt{AttributeError}
In almost every case the most frequently encountered error, \texttt{KeyError}, occurs when solve\_problem() attempts to iterate over the rooms in the graph and inadvertently attempts to access an element of the graph that does not exist. On several occasions, the model responded to this feedback by implementing a try-except block in the definition of solve\_problem(), such as in the following: 

\begin{figure}
    \centering
    \includegraphics[width=\linewidth]{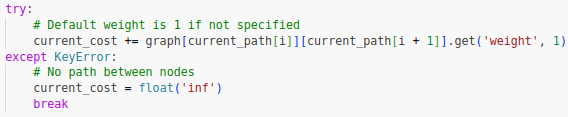}
    \caption{This excerpt from one of the solutions produced by the model exemplifies the purpose of the feedback loop. In this case, the model adopts a brute-force computation to find the least-cost path. }
    \label{fig:enter-label}
    \vspace{-12pt}
\end{figure}


\subsection{Evaluation Summary}
\label{subsec:evalsummary}
Direct prompting and chain-of-thought approaches are capable of generating correct paths in most cases, however their performance declines sharply when measured by the efficiency of the paths they produce. In the least difficult scenario, these methods produce optimal paths 20\% less frequently than the proposed method. \myname{} maintains an efficiency of $\geq96\%$ for all trials, while other methods fall dramatically, demonstrating $\leq20\%$ efficiency in more difficult trials. 
Overall, \myname{} outperforms direct and chain-of-thought approaches in terms of success rate and optimal path planning in all cases. As the number of rooms grows, so does the complexity of the planning problem and the inference time required to plan a path in that environment. While all models perform well in terms of their ability to generate plausible paths, our neuro-symbolic approach significantly outperforms others in the creation of optimal path plans from natural language inputs. Our approach produces paths that are only an average of 1\% longer than the optimal path lengths for the shortest path problem. 


\section{Conclusion}
\label{sec:conclusion}
We propose a prompting framework for path planning that leverages the generalization, abstraction, and code-generating abilities of LLMs and combines them with symbolic verification performed by the Python interpreter. Expanding on previous work in the domains of programmatic prompting and path planning, we establish a feedback loop between the Python interpreter and the LLM that allows the model to correct its mistakes before a final path plan is submitted by the framework. We use partially constructed prompts containing programming instructions for the model that are then combined with the natural language inputs required to solve the problem. The framework then produces a Python program, which is fed to the Python interpreter to return a path in our environment. Our experiments demonstrate that \myname{} improves LLM task performance across multiple metrics. 

 \bibliographystyle{IEEEtran}
 \bibliography{main}

\end{document}